\documentclass{article}
\usepackage{spconf,amsmath,epsfig}
\usepackage{booktabs}
\usepackage[noend]{algpseudocode}
\usepackage{algorithmicx,algorithm}
\usepackage{url}

\let\OLDthebibliography\thebibliography
\renewcommand\thebibliography[1]{
  \OLDthebibliography{#1}
  \setlength{\parskip}{0pt}
  \setlength{\itemsep}{0pt plus 0.3ex}
}

\begin{document}\sloppy

\def\x{{\mathbf x}}
\def\L{{\cal L}}

\title{Invertible ``Mask'' Network for Face Privacy-Preserving}
%
\name{Yang Yang, YiYang Huang, Ming Shi, Kejiang Chen*, Weiming Zhang, Nenghai Yu}
\address{Anhui University, University of Science and Technology of China}

\maketitle

\begin{abstract}
 Face privacy-preserving is one of the hotspots that arises dramatic interests of research. However, the existing face privacy-preserving methods aim at causing the missing of semantic information of face and cannot preserve the reusability of original facial information. To achieve the naturalness of the processed face and the recoverability of the original protected face, this paper proposes face privacy-preserving method based on Invertible ``Mask'' Network (IMN). In IMN, we introduce a Mask-net to generate ``Mask'' face firstly. Then, put the ``Mask'' face onto the protected face and generate the masked face, in which the masked face is indistinguishable from ``Mask'' face. Finally, ``Mask'' face can be put off from the masked face and obtain the recovered face to the authorized users, in which the recovered face is visually indistinguishable from the protected face. The experimental results show that the proposed method can not only effectively protect the privacy of the protected face, but also almost perfectly recover the protected face from the masked face.
 
\end{abstract}
\begin{keywords}
Face privacy-preserving, Invertible ``Mask'' Network, ``Mask'' face
\end{keywords}
\section{Introduction}
\label{sec:intro}

Computer vision technology has been widely used in visual recognition and other tasks. All of these technologies bring great convenience to people's daily life, but also bring huge risks. Large amounts of original photos and videos are uploaded to the cloud or sent to a third party for analysis and recognition tasks, facial information is also included in it. However, facial information is sensitive information that contains a lot of personal information. If not carefully protected, highly sensitive facial information can be easily accessed and illegally used by third parties or malicious attackers. Therefore, the privacy of sensitive information should be protected. Therefore, we need a technology that can ensure the security of facial information while performing conventional computer vision tasks.

 There are some existing face privacy-preserving methods that have been proposed. In previous research, the face privacy-preserving methods are achieved by irreversibly processing the original face using super-pixels~\cite{DBLP:journals/ijmms/House09}, blur ~\cite{DBLP:conf/cikm/ZerrSH12} or low resolution ~\cite{DBLP:journals/tkde/SquicciariniLSW15}. However, these face privacy-preserving methods all aim at causing the missing of semantic information of face and cannot preserve the reusability of the original facial information. With the development of deep learning technology, researchers in the field of face privacy-preserving began to focus on GAN. Maximov et al. proposed CIAGAN~\cite{DBLP:journals/corr/abs-2005-09544}, a model for image and video anonymization based on conditional generative adversarial networks. You et al. proposed a novel reversible face privacy-preserving scheme~\cite{you2021reversible}, which anonymizes the face information and restores the original face information when needed. However, its face anonymization process is adding mosaic on the face which is poor in subjective vision and the protected face is not completely reversible.
 
Through analyzing all of the above-mentioned methods, it can be found that these existing face privacy-preserving methods prevent malicious third parties and legitimate users from accessing the original facial information by retaining only semantic information but permanently destroying the original facial information before uploading, and almost all of these methods cannot fully recover the original face. In fact, due to the obscured face is unnatural and perceptible, it is more likely be noticed by attackers. In addition, in some applications, we hope the obscured face can be imperceptible and the original protected face needs to be recovered. For example, in social platforms, a lot of people like to record their lives and share their photos, they want the photos to look natural without leaking face privacy to unauthorized people, and they also want the original face can be displayed to authorized people (friends or family members, etc). If a criminal is captured in video surveillance, the original face of the criminal also needs to be recovered after privacy preserving.

Therefore, there are two problems that need to be solved. One is how to achieve face privacy-preserving method without arousing suspicion, that is, how to obscure the protected face very naturally; the other is how to achieve reversibility, that is, the obscured image can be recovered to the protected face perfectly. In this paper, we propose IMN for face privacy-preserving. IMN includes three parts: Firstly, generate a ``Mask'' face naturally by Mask-net, which inspirited from SimSwap~\cite{DBLP:journals/corr/abs-2106-06340}; Secondly, put the ``Mask'' face on the protected face and generate the masked face, in which the masked face is visually indistinguishable from the ``Mask''. Finally, put off the ``Mask'' face from the masked face and obtain the recovered face, in which the recovered face is visually indistinguishable from the protected face. It's worth noting that although the visual effect of proposed method is similar to Deepfake technology, however it cannot recovery the original face.
The main contributions of this paper are summarized as follows:
\begin{itemize}
    \item Propose a face privacy-preserving method based on Invertible ``Mask'' Network, which can not only protect the sensitive face, but also almost perfectly recover it.
    \item To achieve considerable visual quality, a face mask-net is introduced to generate the ``Mask''. 
    \item Extensive experiments validate the effectiveness of the proposed Invertible ``Mask'' Network under different evaluation metrics.
\end{itemize}


\section{Method}

\begin{figure}[h]
    \centering
    \includegraphics[width=0.5\textwidth]{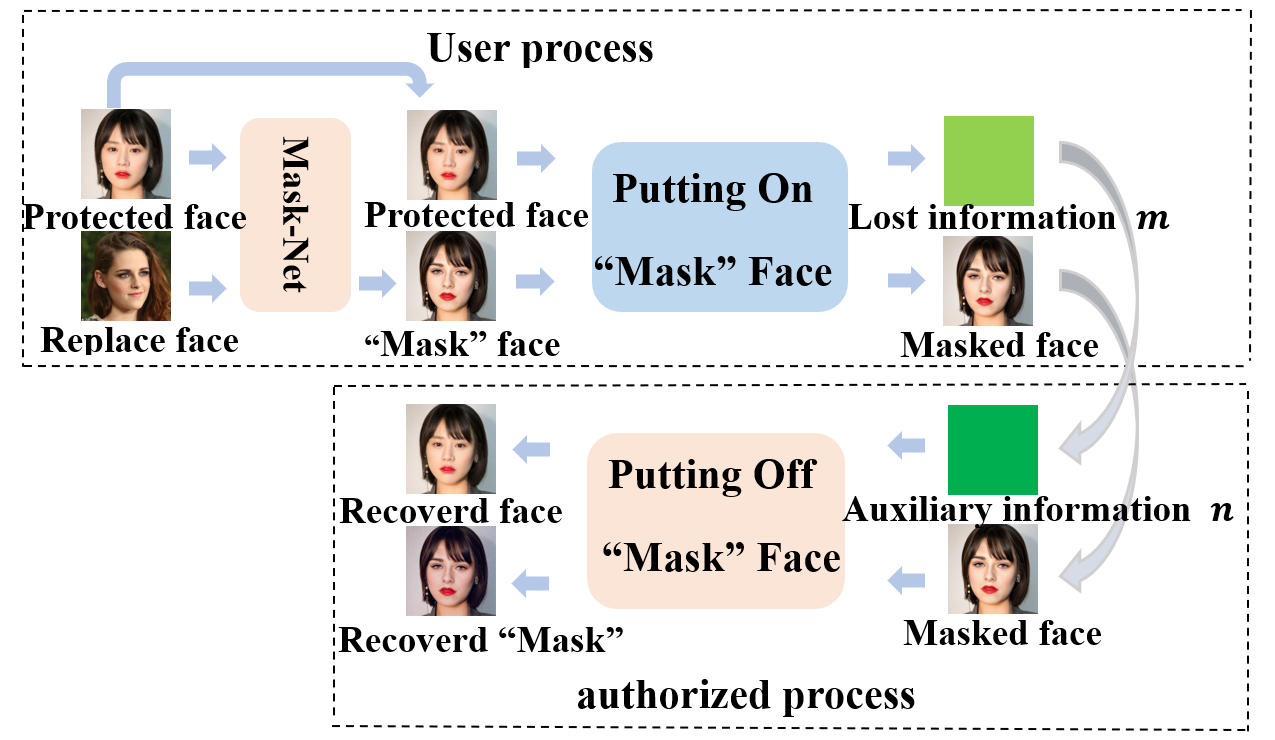}
    \caption{The framework of Invertible “Mask” Network (IMN).}
    \label{detail-frameworkofIMN}
\end{figure}

In this paper, we propose a face privacy-preserving method based on Invertible ``Mask'' Network (IMN). Firstly, according to the protected face and the replaced face, we use Mask-net to generate a ``Mask'' face, which is very natural, because it is highly correlated with the protected face. Secondly, ``Mask'' is put on to the protected face to obtain the masked face, in which the masked face is visually indistinguishable from the ``Mask''. As for authorized users, ``Mask'' face can be put off from the masked face and the protected face is recovered, in which the recovered face is nearly same as the protected face. The overview of the proposed method is shown in Fig.~\ref{detail-frameworkofIMN}.

\begin{table}[h]
\centering
\caption{Summary of notations in this paper}
\begin{tabular}{ c|l }
\hline
\textbf{Notation}                                                                                                              & \textbf{~ ~ ~ ~ ~ ~ ~ ~ ~ ~ ~ ~ ~ ~ ~ ~ ~  ~Description}                                                                                                                                                                                                                                                                                                                                                                   \\
\hline
\begin{tabular}[c]{@{}l@{}}$x_{\text{Protected}}$\\$x_{\text{Replace}}$\\$x_{\text{Mask}}$\\\\$x_{\text{Masked}}$\\$x_{\text{Recovered}}$\\$x_{\text{R-Mask}}$\\$m$\\$n$\\\\
\end{tabular} & \begin{tabular}[c]{@{}l@{}}Protected face: the face to be protected.\\Replace face: the face used for exchanging.\\“Mask” face: the face generated by \\Mask-net.\\Masked face: the face after putting on Mask.\\Recovered face: the face after putting off Mask.\\The recovered ``Mask''.\\The lost information for putting on “Mask”.\\The auxiliary information for putting off~\\“Mask”. ~ ~~\end{tabular} \\
\hline
\end{tabular}

\end{table}

\subsection{Generating ``Mask'' face}\label{sec-masknet}
In order to achieve reversible face privacy-preserving, we generate ``Mask'' face according to the protected face and replace face by using Mask-net. In Mask-net, we adopt SimSwap framework to exchange the protected face with the replaced face ~\cite{DBLP:journals/corr/abs-2106-06340}, while not changing the attributes outside the protected face. Though SimSwap framework generates ``Mask'' face quickly, the detail of it is not precise enough.  Considering that, we then improve the generated face with super-resolution reconstruction and generate the high quality ``Mask'' face.

\begin{figure}[h]
    \centering    \includegraphics[width=0.5\textwidth]{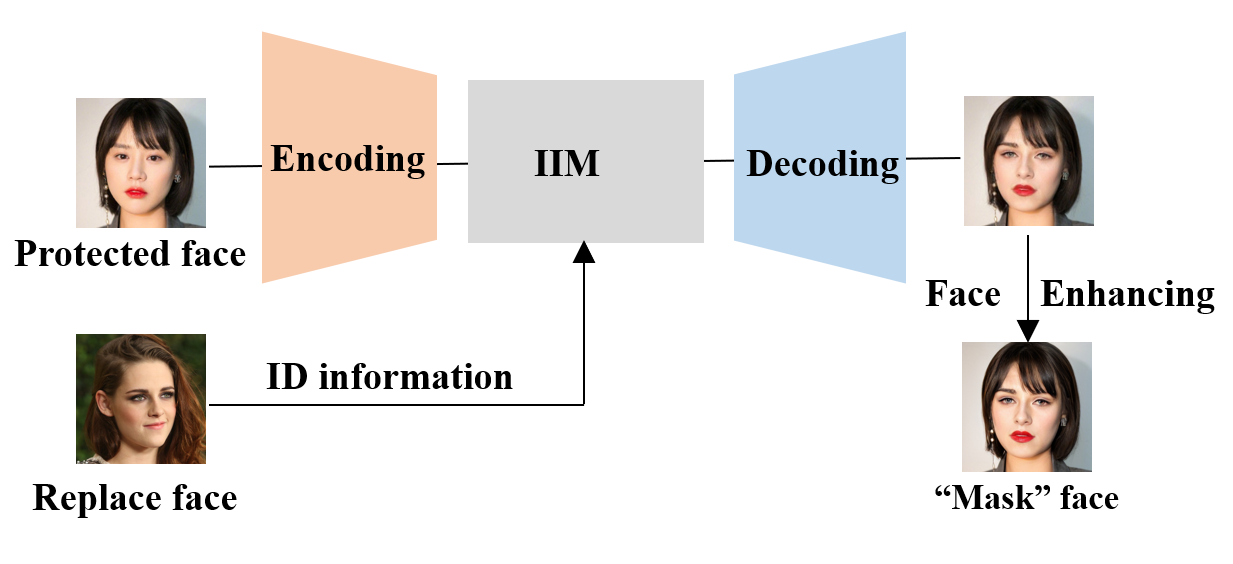}
    \caption{The diagram of Mask-Net.}
    \label{detail-frameworkofMask-Net}
\end{figure}

As shown in Fig.~\ref{detail-frameworkofMask-Net}, the Mask-net consists of four parts, including the Encoding module, the Identity Injection Module (IIM), the Decoding module, Face Enhancing module. The Protected face is encoded in the encoding module and combined with the replaced face in the IIM. Through the decoding part, the primary face is obtained. Finally, the ``Mask'' face is generated by using GAN Prior Embedded Network (GPEN)~\cite{DBLP:journals/corr/abs-2105-06070} to improve the details of the primary face.

\subsection{Putting on the ``Mask'' face} \label{sec-putonmask}

\begin{figure*}[htbp]
    \centering
    \includegraphics[width=0.9\textwidth]{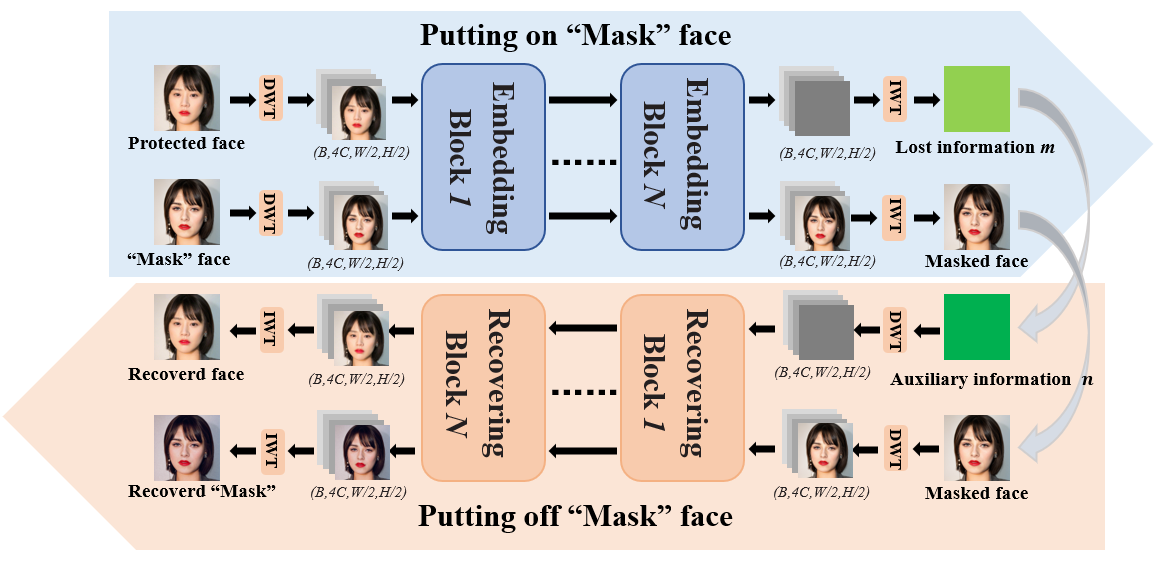}
    \caption{The diagram of putting on and putting off ``Mask'' face.}
    \label{detail-framework}
\end{figure*}

After generating the ``Mask'' face by Mask-net in Section~\ref{sec-masknet}, we introduces how to put the ``Mask'' face on the protected face and generate the masked face. Inspired by the invertible neural network (INN) ~\cite{DBLP:conf/cvpr/LuWZR21}, this paper proposes an embedding architecture to implement the process of putting on the ``Mask'' face and a recovering architecture to implement the process of putting off the ``Mask'' face, respectively, as shown in Fig.~\ref{detail-framework}. The core purpose of putting the protected face on the ``Mask'' face is to embed the protected face $x_{\text{Protected}}$ into the ``Mask'' face $x_{\text{Mask}}$. The algorithm of putting on the ``Mask'' face is presented in Algorithm 1. We summarize the process of putting on the ``Mask'' face as:
\begin{equation}
  \left(x_{\text{Masked}}\ ,m\right)\ =\ f\left(x_{\text{Protected}}\ ,\ x_{\text{Mask}}\right) 
\end{equation}

\begin{algorithm}
	\caption{Putting on ``Mask'' face.}
	\label{Algo1}
	\begin{algorithmic}[1]
		\Require Protected face $x_{\text{Protected}}$ and ``Mask'' face $x_{\text{Mask}}$;
	
		\Ensure the lost information $m$ and Masked face $x_{\text{Masked}}$;
		\State Input Protected face $x_{\text{Protected}}$ and ``Mask'' face $x_{\text{Mask}}$; 
		\State Do the wavelet transform DWT on the Protected face $x_{\text{Protected}}$ and the ``Mask'' face $x_{\text{Mask}}$;
		\State Through the embedding process, the Protected face $x_{\text{Protected}}$ is embedded into the ``Mask'' face $x_{\text{Mask}}$;
		\State Do the wavelet transform IWT on the output of Step3; 
		\State \textbf{return}  the lost information $m$ and Masked face $x_{\text{Masked}}$.

	\end{algorithmic}
	\label{Algorithm 1}
\end{algorithm}

In Algorithm 1, we need to do the wavelet transform in protected face $x_{\text{Protected}}$ and ``Mask'' face $x_{\text{Mask}}$ at first. In general, embedding information in the pixel domain often produces some artifacts, resulting in poor visual effects of the embedded information cover image. On the contrary, embedding information in the frequency domain, especially in the high-frequency region of the image, will course less impact on the visual quality of the image. In summary, we use discrete wavelet transform (DWT) to divide the protected face $x_{\text{Protected}}$  and ``Mask'' face $x_{\text{Mask}}$ into low-frequency and high-frequency wavelet sub-bands, so that the ``Mask'' face $x_{\text{Mask}}$ can hide the protected face $x_{\text{Protected}}$ with less distortion.
Taking into account the complexity of the calculation, we choose Haar wavelet to implement DWT. We define the changes of the size of the feature map through DWT:
\begin{equation}
 (B, C, H, W) \stackrel{D W T}{\rightarrow}\left(B, 4 C, \frac{H}{2}, \frac{W}{2}\right) 
\end{equation}
where $B$ is the batch size, $H$ is the height, $W$ is the width and $C$ is the channel number.
After DWT, the protected face $x_{\text{Protected}}$ is embedded into the ``Mask'' face $x_{\text{Mask}}$. In order to embed the protected face $x_{\text{Protected}}$ into the ``Mask'' face $x_{\text{Mask}}$ reversibly, we use an invertible embedding module ~\cite{DBLP:conf/cvpr/LuWZR21}. There are $N$ embedding blocks with the same architecture in this invertible embedding module. For $i$-th embedding block in this module, the inputs are the $x_{\text{Protected}}^i$ and $x_{\text{Mask}}^i$, and  
the outputs $x_{\text{Protected}}^{i+1}$ and $x_{\text{Mask}}^{i+1}$ are calculated by:
\begin{equation}
 x_{\text {Mask }}^{i+1}=x_{\text {Mask }}^{i}+\varphi\left(x_{\text {Mask }}^{i}\right) 
\end{equation}
\begin{equation}
 x_{\text {Protected }}^{i+1}=x_{\text {Protected }}^{i} \exp \odot\left(\alpha\left(\rho\left(x_{\text {Mask }}^{i+1}\right)\right)\right)+\eta\left(x_{\text {Mask }}^{i+1}\right) 
\end{equation}
where $\alpha$ is a sigmoid function, and $\rho(·)$, $\varphi(·)$, $\eta(·)$ are represented by dense blocks in this paper. The details of different architectures of $\rho(·)$, $\varphi(·)$, and $\eta(·)$ can be referred to~\cite{DBLP:journals/corr/abs-1809-00219}. In the $N$-th embedding block, we do the inverse wavelet transform (IWT) with outputs $x_{\text{Protected}}^N$ and $x_{\text{Mask}}^N$, then obtain the masked face $x_{\text{Masked}}$ and lost information $m$:
\begin{equation}
 \left\{\begin{array}{c}x_{\text {Protected }}^{N} \\ x_{\text {Mask }}^{N}\end{array} \stackrel{I W T}{\rightarrow}\left\{\begin{array}{c}m \\ x_{\text {Masked }}\end{array}\right.\right. 
\end{equation}
 In the process of putting on the ``Mask'' face, it will cause a certain information of the protected face $x_{\text{Protected}}$ to be lost, we define the lost information as $m$. Because of the reversible constraint of IMN, the auxiliary information  $n$ and lost information $m$ obey the same distribution. We randomly sample from a case-agnostic distribution to generate auxiliary information $n$ which is supposed to obey the same distribution as $m$. In the process of putting off the ``Mask'' face, the auxiliary information $n$ is used to help obtain the recovered face $x_{\text{Recovered}}$.

\subsection{Putting off the ``Mask'' face}\label{sec-putoffnmask}
 The invertible architecture of IMN helps put off the ``Mask'' from the masked face $x_{\text{Masked}}$ and obtain recovered face $x_{\text{Recovered}}$ to the authorized users, as shown in Fig.~\ref{detail-framework}. The core purpose of putting off the ``Mask'' face from the masked face $x_{\text{Masked}}$ is to generate recovered face $x_{\text{Recovered}}$. Algorithm 2 depicts the detail of putting off the ``Mask'' face. 
\begin{algorithm}
	\caption{Putting off ``Mask" face.}
	\label{Algo2}
	\begin{algorithmic}[1]
		\Require Masked face $x_{\text{Masked}}$ and auxiliary information $n$;
	
		\Ensure Recovered face $x_{\text{Recovered}}$ and Recovered ``Mask" $x_{\text{R-Mask}}$;
		\State Input masked face $x_{\text{Masked}}$ and auxiliary information $n$; 
		\State Do the wavelet transform DWT on masked face $x_{\text{Masked}}$ and auxiliary information $n$;
		\State Through the recovering process, the recovered face $x_{\text{Recovered}}$ and the recovered ``Mask" face $x_{\text{R-Mask}}$ in the frequency domain are obtained according to masked face $x_{\text{Masked}}$ and auxiliary information $n$;
		\State Do the wavelet transform IWT for the output of Step3; 
		\State \textbf{return}  Recovered face $x_{\text{Recovered}}$ and Recovered ``Mask" $x_{\text{R-Mask}}$.

	\end{algorithmic}
	\label{Algorithm 2}
\end{algorithm}

We summarize the process of putting off the ``Mask'' face as:
\begin{equation}
 \left(x_{\text {Recovered }},x_{\text{R-Mask}}\right)=f\left(x_{\text {Masked }}, n\right) 
\end{equation}

In Algorithm 2, we aim at obtaining the recovered face $x_{\text{Recovered}}$. At first, we input the masked face $x_{\text{Masked}}$ and auxiliary information $n$. In Section~\ref{sec-putonmask}, before we put the ``Mask'' face $x_{\text{Mask}}$ on the protected face $x_{\text{Protected}}$, we do the Haar wavelet transform on them. Here we also need to do Haar wavelet on the masked face $x_{\text{Masked}}$ and auxiliary information $n$ in the invertible process. The process of DWT is same as Section~\ref{sec-putonmask}. After DWT, we obtain the frequency of recovered face and auxiliary information. Then, we design a recovering module which is invertible to embedding module. The structure of this module is similar to the embedding module but with converse directions. There are $N$ recovering blocks with the same architecture in this recovering module. For $i$-th recovering block in this module, the inputs are the masked face $x_{\text{Masked}}^i$ and auxiliary information $n^i$, the outputs $n^{i+1}$ and $x_{\text{Masked}}^{i+1}$ are calculated by:
\begin{equation}
    n^{i+1}=\left(n^{i}-\eta\left(x_{\text {Masked }}^{i}\right)\right) \exp \odot\left(-\alpha\left(\rho\left(x_{\text {Masked }}^{i}\right)\right)\right)
\end{equation}
In the $N$-th recovering block, we also do the IWT to output $n^N$ and $x_{\text{Masked}}^N$ , then obtain the recovered face $x_{\text{Recovered}}$ and  recovered ``Mask" $x_{\text{R-Mask}}$.
\begin{equation}
    \left\{\begin{array}{cc}n^{N} \quad  \\ x_{\text{Masked}}^{N}\end{array} \stackrel{I W T}{\rightarrow} \{\begin{array}{ll}x_{\text {Recovered }}\quad \\ x_{\text{R-Mask}}\end{array}\right. 
\end{equation}
Finally, we obtain the recovered face $x_{\text{Recovered}}$ which is not only visually consistent with the protected face but also is almost same at the pixel level.

\subsection{Loss Function}
The total loss function is composed of three losses: the embedding loss to guarantee the embedding performance, the recovering loss to ensure the recovering performance, and a high-frequency wavelet loss to enhance the security performance.

\subsubsection{Embedding Loss}
The purpose of the embedding process is to embed the protected face $x_{\text{Protected}}$ into the ``Mask'' face $x_{\text{Mask}}$. It is required that the masked face $x_{\text{Masked}}$ is visually identical from the ``Mask'' face $x_{\text{Mask}}$. For this purpose, we define the embedding loss as:
\begin{equation}
    L_{\text{Embedding}}\left(\theta\right)=\sum_{t=1}^{T}{\ell_e\left(x_{\text{Mask}}^{(t)},x_{\text{Masked}}^{(t)}\right)}
\end{equation}
where $x_{\text{Masked}}^{\left(i\right)}$ is equal to $f_\theta\left(\left(x_{\text{Mask}}^{(t)},x_{\text{Recovered}}^{(t)}\right)\right)$, $\theta$ is the parameter in the network, $T$ is the number of training samples and $\ell_e$ is to measure the difference between the masked face $x_{\text{Masked}}$ and the ``Mask'' face $x_{\text{Mask}}$.

\subsubsection{Recovering Loss}

The core purpose of the recovering process is to restore recovered face $x_\text{Recovered}$ from masked face $x_{\text{Masked}}$ and lost information maps. It is required that the recovered face $x_{\text{Recovered}}$ is identical from the protected face $x_{\text{Protected}}$ not only at the visual level but also at the pixel level. For this purpose, we define the recovering loss as:
\begin{equation}
    L_{\text {Recovering }}(\theta)=
    \sum_{t=1}^{T} \mathrm{E}_{\text {maps } \chi}\left[\ell_{R}\left(x_{\text {Protected }}^{(t)}, x_{\text {Recovered }}^{(t)}\right)\right]
\end{equation}
where the recovered face $x_{\text{Recovered}}^{\left(t\right)}$ is equivalent to $f_\theta^{-1}\left(x_{\text{Masked}}^{(t)},n\right)$, $f_\theta^{-1}$ indicates the recovering process, $\chi$ is the Gaussian distribution of map and $\ell_R$ measures the difference between recovered face $x_{\text{Recovered}}$ and protected face $x_{\text{Protected}}$.

\subsubsection{Low-frequency wavelet loss}

The purpose of embedding loss is to be visually indistinguishable between the masked face $x_{\text{Masked}}$ and the ``Mask'' face $x_{\text{Mask}}$. It is known that modifying in the high-frequency domain of the image has little effect on the visual effect. In order to ensure that more information is embedded into the high-frequency sub-band, it is required that the low-frequency sub-band of the masked face $x_{\text{Masked}}$ and the low frequency sub-band of the ``Mask'' face $x_{\text{Mask}}$ are as similar as possible. For this purpose, we define the low-frequency wavelet loss as:
\begin{equation}
    L_{\text{Low-frequency}}(\theta)=
\sum_{t=1}^{T} \ell_{\mathcal{F}}\left(H\left(x_{\text {Masked }}^{(t)}\right)_{L L}, H\left(x_{\text {Mask }}^{(t)}\right)_{L L}\right)
\end{equation}
where ${H()}_{LL}$ represents the extraction of wavelet low frequency sub-bands and $\ell_\mathcal{F}$ indicates the difference between the low-frequency sub-bands of the masked face $x_{\text{Masked}}$ and the ``Mask'' face $x_{\text{Mask}}$.

\subsubsection{Total Loss}
The total loss function $L_{\text{Total}}$ is a weighted sum of embedding loss $L_{\text{Embedding}}$, recovering loss $L_{\text{Recovering}}$ and the low-frequency wavelet loss $L_{\text{Low-frequency}}$:
\begin{equation}
    L_{\text{Total}}=
    \lambda_1L_{\text{Embedding}}+\lambda_2L_{\text{Recovering}}+\lambda_3L_{\text{Low-frequency}}
\end{equation}
where $\lambda_1$, $\lambda_2$ and $\lambda_3$ are weights for balancing different loss functions. We discuss the influence of these parameters in the next section.

\section{Experiments}
\subsection{Experimental Settings}

The experiments are tested on Windows 10 by Pycharm2021 and with Inter(R) Core(TM) i9 5.0 GHz CPU and 64.0 GB RAM. The experiments are carried out on public database, namely, AGE\_ADULTS\cite{G-Lab}. The AGE\_ADULTS database consists of about 10000 images with $1024 \times 1024$ resolution. In this experiment, we randomly choose 800 images as training datasets from AGE\_ADULTS. 

In experiment, four evaluation metrics are used to measure the performance, which include Peak Signal-to-Noise Ratio (PSNR), Structural Similarity Index (SSIM), Root Mean Square Error (RMSE) and Mean Absolute Error (MAE). The larger values of PSNR, SSIM and smaller values of RMSE, MAE indicate higher image quality. We adopt the dense block for $\rho(·)$, $\varphi(·)$ and $\eta(·)$, because it gets the best PSNR results~\cite{DBLP:journals/corr/abs-1809-00219}. In addition, IMN is trained on NVIDIA RTX 3090, and we set the batch size to 16 to make full use of GPU. The learning rate set as 1e-5, weight decay set as 1000 and epochs parameters set as 10000 according to the mainstream setting in~\cite{DBLP:journals/corr/abs-1809-00219}. At last, we set $\lambda_1$, $\lambda_2$ and $\lambda_3$ to $1:3:1$ for the best recovery and discuss the impact of three ratios on the recovery effect in the next subsection.

\subsection{The influence of parameter ${\lambda}_\mathbf{1}$, ${\lambda}_\mathbf{2}$ and ${\lambda}_\mathbf{3}$}

In loss function, the $\lambda_1$, $\lambda_2$ and $\lambda_3$ are weights for balancing different loss functions. Here, we discuss the influence of parameter $\lambda_1$, $\lambda_2$ and $\lambda_3$ in experiment. The parameter of $\lambda_1$ and $\lambda_3$ are related to the masked face $x_{\text{Masked}}$ and the ``Mask'' face $x_{\text{Mask}}$, and make them identical at the visual level. The parameter of $\lambda_2$ is related to the recovered face $x_{\text{Recovered}}$ and the protected face $x_{\text{Protected}}$, and makes them to be indistinguishable not only at the visual level but also at the pixel level. One pursue of this paper is to achieve high quality recovered face $x_{\text{Recovered}}$ to protect protected face meanwhile without reducing the quality of masked face, so we set the quality between the recovered face $x_{\text{Recovered}}$ and the protected face $x_{\text{Protected}}$ as evaluation metric. We explore the empirical parameter on the randomly chosen 20 test images from AGE\_ADULTS database. The results are shown in Table~\ref{Table2}. As we can see, with the increase of parameter $\lambda_2$, the better the experiment performance, which means the closer recovered face $x_{\text{Recovered}}$ to the protected face $x_{\text{Protected}}$ at the pixel level. The ratio $1:3:1$ gets the best recovery result. Therefore, in this paper, we set $1:3:1$ in the following experiments.

\begin{table}[h]
\centering
\caption{ Evaluation results at different $\lambda_1:\lambda_2:\lambda_3$ ratios.}
\begin{tabular}{l|l|l|l|l} 
\toprule
$\lambda_1:\lambda_2:\lambda_3$    & 1:1:1 & 1:2:1 & 1:3:1 & 1:4:1  \\ 
\midrule
PSNR & 38.42 & 46.35 & 47.09 & 46.15  \\ 
SSIM & 0.943 & 0.988 & 0.991 & 0.988  \\ 
RMSE & 9.945 & 1.579 & 1.437 & 1.706  \\ 
MAE  & 2.108 & 0.905 & 0.829 & 0.930  \\
\bottomrule
\end{tabular}
\label{Table2}
\end{table}

\subsection{Subjective and Objective Experimental Results}

\begin{figure}[h]
    \centering
    \includegraphics[width=0.5\textwidth]{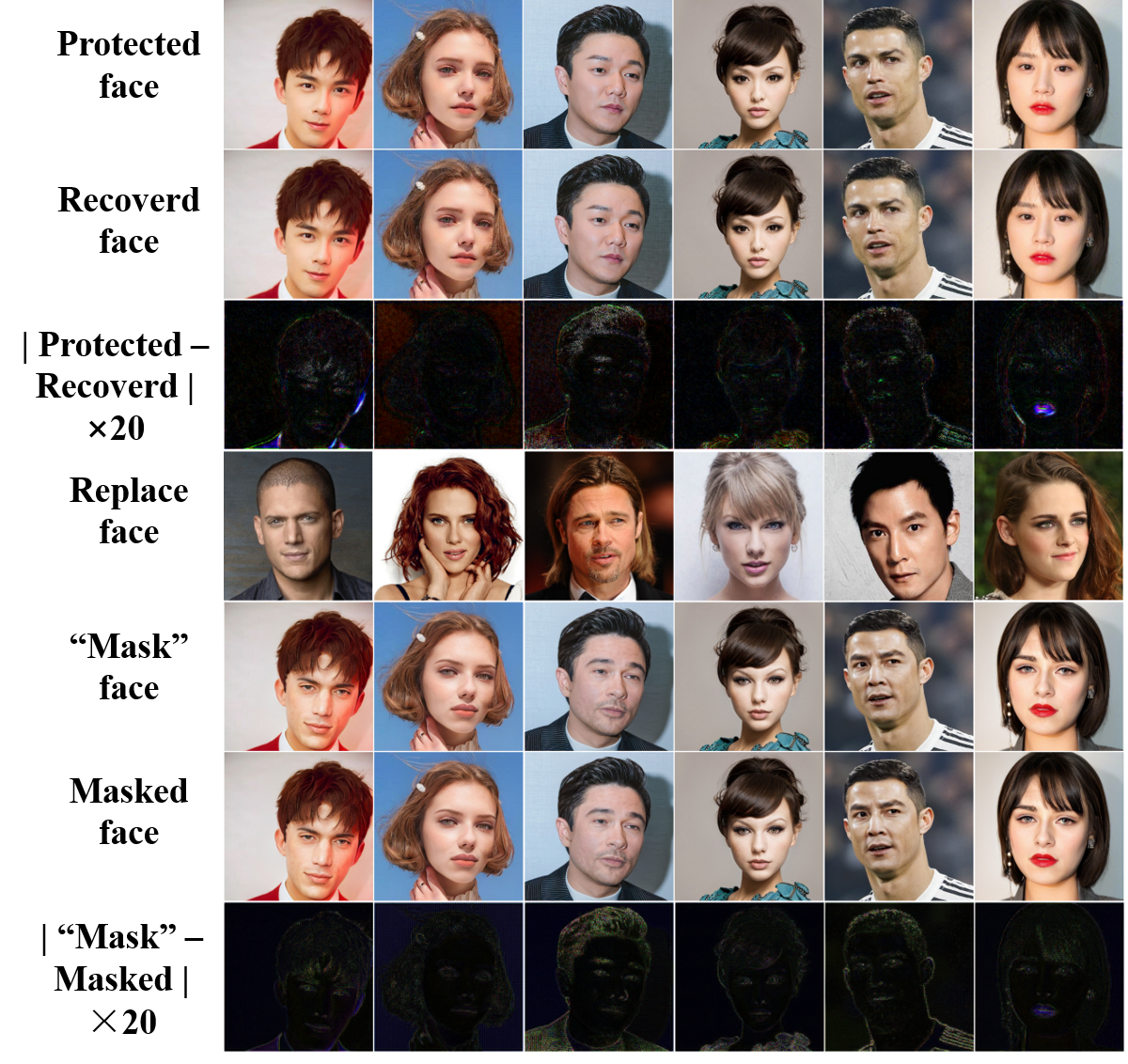}
    \caption{The visualization of faces and differences.}
    \label{fig-resultimg}
\end{figure}

To verify the performance of IMN, we do a series of experiments to discuss the subjective visual effect and objective performance, respectively. The experiments are implemented on images from~\cite{bing}, which is disjoint from our training dataset.
 Fig.~\ref{fig-resultimg} presents six groups subjective experimental results. The first row contains six protected faces, the second row contains six recovered faces for comparison with the first row, the third row contains six difference faces with
$\left|x_{\text{Protected}}-x_{\text{Recovered}}\right|\times20$, the fourth row contains six replace faces which corresponds to the first row, the fifth row contains six ``Mask'' faces, the sixth row contains six masked faces, the seventh row contains six difference faces with $\left|x_{\text{Mask}}-x_{\text{Masked}}\right|\times20$. The results show that the protected faces are well preserved by putting on the ``Mask'' faces, and the features of masked faces are visually different from protected faces. In addition, it can be seen that protected faces and recovered faces are indistinguishable in visual, and ``Mask'' faces are also indistinguishable from masked faces which conceal the protected faces completely. These faces all look natural. In addition, due to the Mask-net containing super-resolution reconstruction part, the quality of ``Mask'' faces is enhanced when compared with protected faces.

In addition to Fig.~\ref{fig-resultimg}, we also use PSNR, SSIM metrics to present experimental results from objective data.  Table~\ref{Tabel3} presents the PSNR, SSIM result values between the recovered face $x_{\text{Recovered}}$ and the protected face $x_{\text{Protected}}$, and between the ``Mask'' face $x_{\text{Mask}}$ and the masked faces $x_{\text{Masked}}$, respectively. The results show that the PSNR and SSIM between $x_{\text{Recovered}}$ and $x_{\text{Protected}}$ can reach 49.46dB and 0.997, respectively, meanwhile the PSNR and SSIM between $x_{\text{Mask}}$ and $x_{\text{Masked}}$ can reach 50.16dB and 0.997, respectively. The results indicate that these pairs of images are almost identical.

\begin{table}
\small
\centering
\caption{The objective results which correspond to Fig. 4.}
\resizebox{0.48\textwidth}{15mm}{
\begin{tabular}{l|c|c|c|c|c|c} 
\toprule
Column                    & 1     & 2     & 3     & 4     & 5     & 6      \\ 
\midrule

\begin{tabular}[c]{@{}l@{}}PSNR\\($x_\text{{Protected}}$,$x_\text{{Recoverd}}$)\end{tabular} & 45.37 & 49.46 & 40.32 & 47.38 & 42.53 & 47.51  \\
\begin{tabular}[c]{@{}l@{}}SSIM\\($x_\text{{Protected}}$,$x_\text{{Recoverd}}$)\end{tabular} & 0.993 & 0.997 & 0.990 & 0.996 & 0.991 & 0.995  \\
\begin{tabular}[c]{@{}l@{}}PSNR\\($x_\text{{Mask}}$,$x_\text{{Masked}}$)\end{tabular} & 50.16 & 47.01 & 43.72 & 47.59 & 45.21 & 49.27  \\
\begin{tabular}[c]{@{}l@{}}SSIM\\($x_\text{{Mask}}$,$x_\text{{Masked}}$)\end{tabular} & 0.997 & 0.990 & 0.992 & 0.996 & 0.992 & 0.996  \\
\bottomrule
\end{tabular} }

\label{Tabel3}
\end{table}

You et al.~\cite{you2021reversible} proposed a reversible mosaic transform for privacy preserving. Obviously, the visual quality of the masked image is far worse ours. Therefore, we only compare the quality of the recovered face. The experiments are carried out on the same dataset used in~\cite{you2021reversible}. As shown in Table~\ref{Tabel5}, the proposed method has better recovering performance than You et al.'s method~\cite{you2021reversible}. The underlying reason is that the invertible network architect performs far better than the autoencoder.

\begin{table}[h]
\centering
\caption{The comparison results between You et al.'s method~\cite{you2021reversible} and proposed method.}
\begin{tabular}{c|c|c|c|c} 
\hline
Methods        & PSNR  & SSIM  & RMSE  & MAE   \\ 
\hline
You et al.~\cite{you2021reversible} & 36.67 & 0.988 & 14.72 & 2.74  \\
proposed method           & 52.02 & 0.997 & 0.441 & 0.45  \\
\hline
\end{tabular}

\label{Tabel5}
\end{table}

\section{Conclusion}
In this paper, we propose a face privacy-preserving method based on Invertible ``Mask'' Network (IMN). Firstly, generate a high quality ``Mask'' face by Mask-net; then put the ``Mask'' face on the protected face and generate the masked face; finally, put the ``Mask'' face off from the masked face and recover the recovered face. The experimental results show that the proposed method is effective in protecting sensitive faces. The features of masked faces are different from protected faces in visual, and the protected face can be recovered almost perfectly by the authorized users.


\bibliographystyle{IEEEbib}
\bibliography{icme2022template}

\end{document}